\documentclass[preprint,12pt]{elsarticle}
\makeatletter
\def\ps@pprintTitle{%
 \let\@oddhead\@empty
 \let\@evenhead\@empty
 \def\@oddfoot{\centerline{\thepage}}%
 \let\@evenfoot\@oddfoot}
\makeatother

\usepackage{graphicx}
\usepackage{amssymb}
\usepackage{subfig}
\usepackage[margin=1.2in,footskip=0.25in]{geometry}
\usepackage{adjustbox}
\usepackage{amsmath}
\usepackage{multirow}

\begin{document}

\begin{frontmatter}


\title{Metaphor Detection using Deep Contextualized Word Embeddings}


\author[1]{Shashwat Aggarwal}
\address[1]{University Of Delhi}
\ead{shashwata.co@nsit.net.in}
\author[2]{Ramesh Singh}
\address[2]{National Informatics Center}
\ead{rsingh@nic.in}

\begin{abstract}
Metaphors are ubiquitous in natural language, and their detection plays an essential role in many natural language processing tasks, such as language understanding, sentiment analysis, etc. Most existing approaches for metaphor detection rely on complex, hand-crafted and fine-tuned feature pipelines, which greatly limit their applicability. In this work, we present an end-to-end method composed of deep contextualized word embeddings, bidirectional LSTMs and multi-head attention mechanism to address the task of automatic metaphor detection. Our method, unlike many other existing approaches, requires only the raw text sequences as input features to detect the metaphoricity of a phrase. We compare the performance of our method against the existing baselines on two benchmark datasets, TroFi, and MOH-X respectively. Experimental evaluations confirm the effectiveness of our approach.
\end{abstract}


\end{frontmatter}


\section{Introduction}
\label{S:1}

A metaphor is a figurative form of expression that compares a word or a phrase to an object or an action to which it is not literally applicable but helps explain an idea or suggest a likeness or analogy between them. Metaphors have been used extensively in all types of literature and writings, especially in poetry and songs to communicate complex feelings, emotions, and visuals present in the text to readers effectively. Metaphors are ubiquitous in natural language and help in structuring our understanding of the world even without our conscious realization of its presence~\cite{lakoff1980}. Given the prevalence and significance of metaphorical language, effective detection of metaphors plays an essential role in many natural language processing applications, for example, language understanding, information extraction, sentiment analysis, etc.

However, automated detection of metaphorical phrases is a difficult problem primarily due to three reasons. First, there is a subjective component involved: the metaphoricity of expression may vary across humans. Second, metaphors can be domain and context dependent. And third, there is a lack of annotated data, which is required to train supervised machine learning algorithms to facilitate automated detection accurately. 

Most of the previous approaches for detection of metaphorical phrases, have either relied on manual and lexical detection~\cite{mason2004,krishnakumaran2007,turney2011} which requires heavily handcrafted features built from linguistic resources, that are costly to obtain and greatly limits their applicability or have used supervised machine learning based algorithms~\cite{tsvetkov2013,shutova2016,rei2017} with limited forms of linguistic context, for example using only the subject verb objects triplets (e.g. \textit{cat eat fish}). Although these techniques automate the detection of metaphorical phrases, however, the prediction accuracies are not as good as the prediction accuracies of these techniques in other text classification tasks.

Inspired by recent works in the field of NLP and transfer learning, in this paper, we present an end-to-end method composed of deep contextualized word embeddings, bidirectional LSTMs and multi-head attention mechanism to address some of the limitations aforementioned. Our method is notable in the sense that unlike many existing approaches, it requires only the raw text sequences as input and does not depend on any complex or fine-tuned feature pipelines.

\section{Literature Survey}
\label{S:2}

There has been significant work done in automatic detection and discovery of metaphorical phrases in natural language ranging from the traditional rule-based methods which rely on task-specific hand-coded lexical resources to the recent statistical machine learning models to identify metaphors. 

One of the first attempts to detect metaphorical phrases automatically in the text was by~\cite{fass1991}. Their system called \textit{met*} could discriminate between literalness, metonymy, and metaphor in the underlying text.~\cite{birke2006} proposed a sentence clustering approach for non-literal language recognition using a  similarity-based word sense disambiguation method.  However, their approach focused only on metaphors expressed by a verb. ~\cite{gedigian2006} trained a maximum entropy classifier to discriminate between literal and metaphorical usage of a phrase. Their method got a high accuracy of 95.12\%. However, the majority of the verbs present in their dataset were used metaphorically, thus making the task notably easier. ~\cite{krishnakumaran2007} used hyponymy (IS-A) relation in WordNet~\cite{wordnet} and word bigram counts to predict metaphors in sentences. However, the word bigram counts lose a great deal of information over the verb-noun pairs. Also, they do not deal with literal sentences.

~\cite{shutova2010} used statistical learning methods to identify metaphors automatically. They start with a small seed of manually annotated metaphorical expressions. The system generates a large number of metaphors of similar syntactic structure from a corpus. Other approaches using statistical learning methods to identify metaphors automatically include~\cite{turney2011}, that use concreteness and abstractness to detect metaphors,~\cite{heintz2013}, that apply a Latent Dirichlet Allocation (LDA) based topic modeling method for automatic extraction of linguistic metaphors,~\cite{tsvetkov2014} that construct a metaphor detection system using a random forest classifier with conceptual semantic features such as abstractness, imageability, and semantic supersenses, etc. Most of these existing approaches use a variety of features which rely on external lexical, syntactic, or semantic linguistic resources to train classification models for metaphor detection which severely limits their applicability.  

Recently, there has been an introduction of several deep learning based approaches to tackle the problem of metaphor detection. Several methods such as~\cite{dinh2016, koper2017, rei2017} train word embeddings to identify and detect metaphors and have shown gains on various benchmarks. ~\cite{sun2017} used bidirectional LSTMs with raw text, SVO and dependency subsequences as input features. The 2018 VUA Metaphor Detection Shared Task has also introduced several LSTM and CRF based models~\cite{mosolova,wu2018,swarnkar2018,bizzoni2018} which are augmented by linguistic features such as POS tags, lemmas, verb clusters, unigrams, WordNet, etc.

Furthermore, there have been works such as ~\cite{gao2018, mu2019} which have employed the recently proposed contextualized language representation models to detect metaphors. However, the work in this direction is limited. Also, a recently proposed popular multi-head attention mechanism~\cite{vaswani2017attention}, which has been used extensively in domains such as speech processing or machine translation in the recent past has also not been tried for the task of metaphor detection yet.

Based on the lines of some of these recent works, in this paper, we combine the contextualized language representation models, ELMo~\cite{elmo} and BERT~\cite{bert} with multi-head attention mechanism and bidirectional LSTMs to detect metaphorical phrases automatically. We compare our proposed approach with existing baselines for the task of metaphor detection and classification. 

The rest of the paper is organized as follows: In section 3, we discuss our proposed approach and in section 4 \& 5, we describe the datasets we use and the experiments we perform on them. 

\section{Proposed Approach}
\label{S:3}

We present an end-to-end method composed of deep contextualized word embeddings, bidirectional LSTMs and multi-head attention mechanism to address the task of automatic metaphor detection and classification. Our method takes the raw text sequence as input and does not depend on any complex, hand-crafted or fine-tuned feature pipelines.

\begin{figure}[!t]
  \includegraphics[scale=0.82]{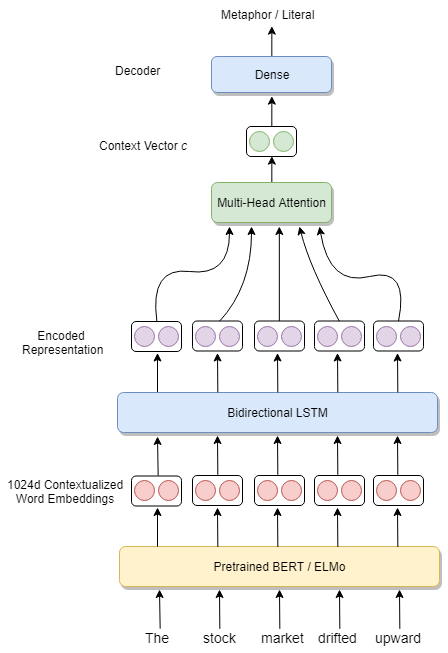}
  \caption{Proposed Model for metaphor detection from raw text sequences.}
  \label{fig:model}
\end{figure}

\subsection{Contextualized Word Representations}

Natural language processing (NLP) is a diversified field consisting of a wide variety of tasks such as text classification, named entity recognition, question answering, etc. However, most of these tasks contain limited datasets with a few thousand or a few hundred thousand human-labeled training examples. This shortage of training data severely affects the performance of modern deep learning-based NLP systems, which proffer benefits when trained from much larger amounts of data. 

Recently, there have been techniques such as~\cite{elmo,bert}, that address this limitation by pretraining language representation models on enormous amount of unannotated text and then transferring the learned representation to a downstream model of interest as contextualized word embeddings. These contextualized word representations have resulted in substantial gains in accuracy improvements in numerous NLP tasks as compared to training on these tasks from scratch. Following recent work, we also use these contextualized word embeddings for our task of metaphor detection.

\subsection{Proposed Model}

Figure~\ref{fig:model} shows the architecture of our proposed method for the task of metaphor classification. Each token \(t_i\) of the input text is encoded by the contextualized word embeddings, ELMo or BERT, to obtain the corresponding contextualized word representations. The contextualized representations are feed as input to the bidirectional LSTM encoder network. The encoder network outputs an encoded representation across every timestep of the input sequence of text.

We add a multi-head attention mechanism on top of the encoder network to obtain a context vector c  representing the weighted sum of all the BiLSTM output states as depicted in equation \ref{eq1},  where \(h\) is the number of attention heads used, \(x_i\) corresponds to the hidden representations from the encoder network, \(a_i^j\) is the attention weight computed for token \(t_i\) from attention head \(j\) and \(c_j\) is the context vector obtained from attention head \(j\) respectively.

\begin{equation}
    \label{eq1}
    \begin{split}
    &a_i^j = Softmax_i(W_{a^j}x_i + b_{a^j}), \\
    &c_j = \sum_{i=1}^{n}a_i^jx_i, \\
    &c = W_o[c_1;c_2;...;c_h] + b, \\ \\
    \end{split}
\end{equation}

Finally, the context vector \(c\) is feed to a dense layer decoder to predict the metaphoricity of the input phrase.

\section{Experiments}
\label{S:4}

\subsection{Datasets}

We evaluate our method on two benchmark datasets, TroFi~\cite{birke2006} and MOH-X ~\cite{shutova2016} respectively. In Table~\ref{data-table}, we report a summary of each of the dataset used in our study. The TroFi dataset consists of literal and nonliteral usage of 50 English verbs drawn from the 1987-89 Wall Street Journal (WSJ) Corpus. The MOH-X dataset, on the other hand, is an adaptation of the MOH dataset  ~\cite{mohammad2016} which consists of simple and concise sentences derived from various news articles.

\subsection{Implementation Details}

We use \(1024d\) ELMo and BERT vectors. The LSTM module has a \(256d\) hidden state. We use \(4\) attention heads for the multi-head attention mechanism. We train the network using Adam optimizer with learning rate \(lr= 0.00003, \beta1 = 0.9, \beta2 = 0.999\) and  with  a  batch  size of \(16\). The network is trained for \(50\) epochs.

\subsection{Baselines}

We compare the performance of our method with four existing baselines. The first baseline is a simple lexical baseline that classifies a phrase or token as metaphorical if that phrase or token is annotated metaphorically more frequently than as literally in the training set. The other baselines include a logistic regression classifier similar to one employed by ~\cite{klebanov2016}, a neural similarity network with skip-gram word embeddings ~\cite{rei2017}, and a BiLSTM based model combined with ELMo embeddings proposed in ~\cite{gao2018}.

\subsection{Evaluation Metrics}

To evaluate the performance of our method with the baselines, we compute the precision, recall, F1 measure for the metaphor class and the overall accuracy. We perform \(10\) fold cross-validation following prior work.

\begin{table}[t]
\centering
\caption{\label{data-table} Metaphor Detection dataset statistics. \% Metaphor refers to sentence-level percentage. }
\begin{tabular}{|c|c|c|c|}
\hline
\textbf{Dataset} & \textbf{\# Examples} & \textbf{\% Metaphors} & \textbf{\# Unique Verbs}  \\ \hline
TroFi            & 3,737                & 43\%                  & 50                        \\ \hline
MOH-X            & 647                  & 49\%                  & 214                       \\ \hline
\end{tabular}
\end{table}

\section{Results}
\label{S:5}

In Table~\ref{result-model-table}, we show the performance of all the baselines and our proposed method on the task of metaphor detection and classification for two benchmark datasets (MOH-X and TroFi). Both of our proposed methods outperform the existing baselines on both the datasets using only the raw text as input features. The model with BERT performs better as compared to the model with ELMo embeddings on both the datasets. Contextualized word embeddings and multi-head attention mechanism improve the performance of both models by a large margin. 

\begin{figure}[!t]
  \centering
  \includegraphics[width=\linewidth]{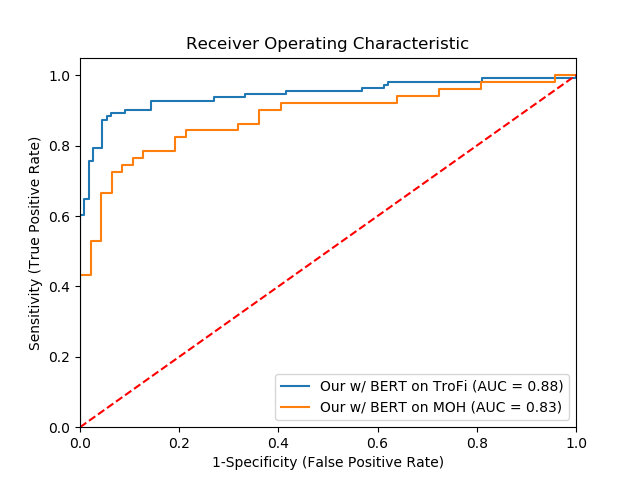}
  \caption{ROC Curve showing the AUC value for the task of metaphor detection for the best model.}
  \label{fig:roc}
\end{figure}

In addition to the performance comparison of our methods with the existing baselines, we also show the receiver operating characteristic curve in Figure~\ref{fig:roc} showing the AUC values for the metaphor detection task and illustrating the diagnostic ability of the best method (ours w/ BERT) as its discrimination threshold is varied. 

\begin{table}[!t]
\centering
\caption{\label{result-model-table} Comparison of our proposed model with baselines on metaphor detection and classification task.}
\begin{tabular}{|c|c|c|c|c|c|c|c|c|}
\hline
\multirow{2}{*}{\textbf{Model}} & \multicolumn{4}{c|}{\textbf{TroFi}}                    & \multicolumn{4}{c|}{\textbf{MOH-X}}                      \\ \cline{2-9} 
                        & \textbf{P}    & \textbf{R}    & \textbf{F1}      & \textbf{Acc}      & \textbf{P}         & \textbf{R}         & \textbf{F1} & \textbf{Acc}      \\ \hline
                        
Lexical Baseline        &72.4           &55.7           &62.9       &71.4       &39.1           &26.7           &31.3       &43.6           \\ \hline
Log. Regression         &70.7           &71.4           &70.3       &72.7       &68.7           &66.2           &67.4       &73.6           \\ \hline
Rei (2017)              &-              &-              &-          &-          &73.6           &76.1  &74.2       &74.8           \\ \hline
Gao (2018) - CLS        &68.7           &74.6           &72.0       &73.7       &75.3           &\textbf{84.3}           &79.1       &78.5           \\ \hline
Gao (2018) - SEQ        &70.7           &71.6           &71.1       &74.6       &79.1           &73.5           &75.6       &77.2            \\ \hline

\textbf{Ours + ELMo}   & 82.7 & \textbf{85.6} & 82.7 & 83.1 & 80.4 & 73.7 & 75.9 & 78.1 \\ \hline
\textbf{Ours + BERT}   & \textbf{85.3} & 84.9 & \textbf{83.2} & \textbf{85.8} & \textbf{90.7} & 74.8 & \textbf{79.8} & \textbf{80.7} \\ \hline

\end{tabular}
\end{table}

Further, In Table~\ref{result-example1-table} \&~\ref{result-example2-table}, we report a set of examples from the TroFi and MOH-X developement sets along with the gold annotation label, the predicted label from both of our methods (with ELMo and with BERT) and the corresponding prediction scores assigned.  The system confidently detects most of the metaporical and literal phrases such as \textit{"make the people's \textbf{hearts glow}."}, \textit{"he \textbf{pasted} his opponent."}  and respectively. However, the predicted output disagrees with the ground truth annotations for some phrases such as \textit{"the white house sits on pennsylvania avenue."}  and \textit{"bicycle suffered major damage."} respectively. Majority of errors committed by our method are in detection of a particular type of metaphor called personification. The metaphoricity of phrases such as \textit{"the white house sits on pennsylvania avenue."} are largely dependant on a few words like \textit{"white house"} which if replaced by the name of some person changes the metaphoricity of the phrase. The network occasionally gets confused in discriminating between different entities like people, things, etc. Given additional information along with raw text as input to the network, for example the part-of-speech or named-entity tags, the performance of the network could be improved.

Finally, we report some example phrases such as \textit{"vesuvius erupts once in a while."} or \textit{"the old man was sweeping the floor."} for which the model with BERT embeddings correctly detects the metaphoricity of the phrase, while the model with ELMo fails to do so.

\begin{table*}[!ht]
\centering
\caption{\label{result-example1-table}  Examples from the TroFi development set, along with the gold label, predicted label, and the predicted score from our method with ELMo and BERT.}
\begin{adjustbox}{width=1\textwidth}
\begin{tabular}{|c|c|c|c|c|c|}
\hline
\multirow{2}{*}{Input Phrase} & \multirow{2}{*}{Gold} & \multicolumn{2}{c|}{Predicted} & \multicolumn{2}{c|}{Score (for Metaphor Class)} \\ \cline{3-6} 
                              &                       & w/ ELMo        & w/ BERT       & w/ ElMo      & w/ BERT     \\ \hline
                              \textit{"make the people's hearts glow ."}                       & 1               &1           &1    & 0.995             &0.999             \\ \hline
            \textit{"she leaned over the banister"}                  & 0                      & 0               &0               &0.001              &0.001             \\ \hline
                       \textit{"he pasted his opponent."}       &1                       & 1               & 1              & 0.999             & 0.999             \\ \hline
               \textit{"vesuvius erupts once in a while."}               &0                       & 1               & 0             & 0.993             &  0.001           \\ \hline
         \textit{"the white house sits on pennsylvania avenue."}                     &1                       &0                &0               &0.001              &0.001             \\ \hline
\end{tabular}
\end{adjustbox}
\end{table*}

\begin{table*}[!ht]
\centering
\caption{\label{result-example2-table}  Examples from the MOH-X development set, along with the gold label, predicted label, and the predicted score from our method with ELMo and BERT.}
\begin{adjustbox}{width=1\textwidth}
\begin{tabular}{|c|c|c|c|c|c|}
\hline
\multirow{2}{*}{Input Phrase} & \multirow{2}{*}{Gold} & \multicolumn{2}{c|}{Predicted} & \multicolumn{2}{c|}{Score (for Metaphor Class)} \\ \cline{3-6} 
                              &                       & w/ ELMo        & w/ BERT       & w/ ElMo      & w/ BERT     \\ \hline
                      \textit{"this one speech could sink his candidacy."}        &  1                     &  1              & 1               & 0.999             &  0.999           \\ \hline
        \textit{"attach a drain hose to the radiator drain."}                    & 0                      & 0                &  0             &  0.001            & 0.001            \\ \hline
            \textit{ "the old man was sweeping the floor." }                &   0                    &  1               &  0             &    0.837          &   0.001          \\ \hline
      \textit{ "the object then moved slowly away."}                       &  0                     &  1               &  0             &  0.999            &   0.474          \\ \hline
               \textit{"bicycle suffered major damage."}               & 1                      & 0               &  0             & 0.062             &    0.001         \\ \hline                             
\end{tabular}
\end{adjustbox}
\end{table*}

\section{Conclusion}
\label{S:6}

In this work, we presented an end-to-end method composed of deep contextualized word embeddings, bidirectional LSTMs, and multi-head attention mechanism to address the task of automatic metaphor detection and classification. Our method requires only the raw text sequences as input and does not depend on any complex or fine-tuned feature pipelines. Our method established new state-of-the-art on both the datasets for metaphor detection.






\bibliographystyle{elsarticle-num-names}
\bibliography{main.bib}







\end{document}